\documentclass{article}
\usepackage{longtable}
\usepackage{graphicx}
\usepackage{PRIMEarxiv}
\usepackage{array}
\usepackage[utf8]{inputenc} % allow utf-8 input
\usepackage[T1]{fontenc}    % use 8-bit T1 fonts
\usepackage{hyperref}       % hyperlinks
\usepackage{url}            % simple URL typesetting
\usepackage{booktabs}       % professional-quality tables
\usepackage{amsfonts}       % blackboard math symbols
\usepackage{nicefrac}       % compact symbols for 1/2, etc.
\usepackage{microtype}      % microtypography
\usepackage{lipsum}
\usepackage{fancyhdr}       % header
\usepackage{graphicx}       % graphics
\graphicspath{{media/}}     % organize your images and other figures under media/ folder
\usepackage{parskip} 
\usepackage{tabularray}
\title{Reinforcement Learning for Quadrupedal Locomotion: Current Advancements and Future Perspectives
}

\author{
  Maurya Gurram \\
  Computer Science and Engineering\\
  Vellore Institute of Technology,AP\\
  \texttt{maurya.19bcd7178@vitapalum.ac.in} \\
   \And
  Prakash Kumar Uttam \\
  Electricial, Electronics and Computer Sciences\\
  Indian Institute of Science,Bangalore\\
  \texttt{prakashuttam@iisc.ac.in} \\
   \And 
   Dr. Shantipal S. Ohol\\
   Associate Professor\\
   COEP, Pune\\
   sso.mech@coeptech.ac.in
}

\begin{document}
\maketitle

\begin{abstract}
In recent years, reinforcement learning (RL) based quadrupedal locomotion control has emerged as an extensively researched field, driven by the potential advantages of autonomous learning and adaptation compared to traditional control methods. This paper provides a comprehensive study of the latest research in applying RL techniques to develop locomotion controllers for quadrupedal robots. 
We present a detailed overview of the core concepts, methodologies, and key advancements in RL-based locomotion controllers, including learning algorithms, training curricula, reward formulations, and simulation-to-real transfer techniques. The study covers both gait-bound and gait-free approaches, highlighting their respective strengths and limitations. Additionally, we discuss the integration of these controllers with robotic hardware and the role of sensor feedback in enabling adaptive behavior. The paper also outlines future research directions, such as incorporating exteroceptive sensing, combining model-based and model-free techniques, and developing online learning capabilities.
Our study aims to provide researchers and practitioners with a comprehensive understanding of the state-of-the-art in RL-based locomotion controllers, enabling them to build upon existing work and explore novel solutions for enhancing the mobility and adaptability of quadrupedal robots in real-world environments.
\end{abstract}

% keywords can be removed

\section{Introduction}
Significant progress has been made in the field of legged
locomotion recently in creating robotic systems that can
navigate diverse terrains. To achieve locomotion capabilities
that can tackle diverse and complex terrains, it is crucial to
design controllers that can adapt in real-time to changing
environmental conditions. With reinforcement learning,
It has become possible to design robust controllers that can
withstand frequent changes in terrain and environmental
dynamics. RL-based legged locomotion controllers have
shown comparatively better results than classical controllers
in diverse terrains. By leveraging this approach, robots can
autonomously adapt their locomotion strategies, including leg
coordination, gait selection, and joint control to effectively
navigate through a wide range of terrain. This study is an
attempt to provide an overall understanding of the end-to-end
development and deployment of RL-based controllers on real
quadrupeds. It is also an attempt to highlight certain aspects of
different approaches taken by researchers and scientists in the
past that have enhanced these controllers in various contexts.
The study will consider various learning algorithms.
employed, representations of the simulated environment,
various tools used and training architectures. \newline
\newline
The quadrupedal robot industry has been experiencing substantial growth and is projected to maintain a robust upward trajectory in the coming years. According to market analysis, the global quadruped robot market was estimated to be worth around 1453.6 million dollars US in 2023. However, this market value is expected to witness a significant surge, reaching approximately 4454.9 million dollars US by the year 2030. This growth is anticipated to occur at a remarkable compound annual growth rate (CAGR) of 17.3 percent during the forecast period spanning from 2024 to 2030. These figures highlight the increasing demand and widespread adoption of quadrupedal robotic systems across various sectors and applications. \textbf{The motivation for this study stems from the growing interest in RL-based locomotion controllers and their potential to enhance the mobility and adaptability of quadrupedal robots. As researchers explore novel techniques and architectures, it is crucial to consolidate the existing knowledge and identify promising directions for future research. This study aims to provide a comprehensive understanding of the key concepts, methodologies, and advancements in the field, enabling researchers and practitioners to build upon existing work and explore innovative solutions.}\newline
\newline
The reward and punishment architecture of RL has proven
to be highly appropriate for robotic tasks like qudrupedal
locomotion, humanoid gait planning, and single and dual robotic
arm manipulation. As such tasks are dynamic in nature, RL
Algorithms are a perfect fit as they work towards finding the
right set of actions for such tasks. 
Classical controllers fail to model specific environmental
parameters, as a human designer is only capable of pre-
programming what is in the ‘imaginable range’, this does
not incorporate challenging dynamic terrains. Variations in friction,
terrain height, softness, and dynamics make it challenging.
for the robot to adapt to these abrupt shifts using traditional
control paradigms. Control-based methods have been a highly
researched field in the past decade; with the emergence of
MPC (model predictive control) [1–6] dynamic locomotion
in diverse terrains has become feasible to a certain extent.
Reinforcement learning has proven to be successful
alternative to these control models that tend to fail in diverse
and dynamic terrain. An RL policy can be trained to optimize
these control models to improve performance. The problem
of attaining higher linear speeds while maintaining stability
and different gait parameters seemed impossible, but with the
advent of RL-based locomotion policies, this can be easily

\section{Introduction to RL policies for legged locomotion}
\label{sec:headings}
Reinforcement learning (RL) has emerged as a promising framework for developing control policies that can enable robust legged locomotion in robots. By modeling the robot's physical parameters, such as gravitational forces, orientation, velocity, and joint configurations, as the state representation, RL algorithms can learn to map these states to appropriate actions, expressed as target joint positions. These target positions are then converted into the requisite joint torques by the low-level controllers, thereby actuating the robot's limbs.
Currently, we can replicate environmental
parameters like terrain types and friction to train the
RL-model in simulation before deployment on the real robot,
unlike Tuomas Haarnoja [7], where they train their model directly
in the real world. Many terrain types can be simulated
in an environment like IssacGym [8] or RaiSim along with
the robot’s physical properties expressed as a Unified Robot Description Format (URDF) file. The RL controller is first trained in the simulated environment and then we apply
Sim-To-Real [31] transfer methods to deploy the trained policy on
real robots. Once the policy is optimized in the simulated environment,
we can assume that the deployed policy will perform
in a similar manner, given that the task specified resembles the
task it is trained on. Constant efforts are being made to reduce
the gap between Sim-To-Real transfer, as these conditions \begin{figure}
\centering
  \includegraphics[scale=0.75]{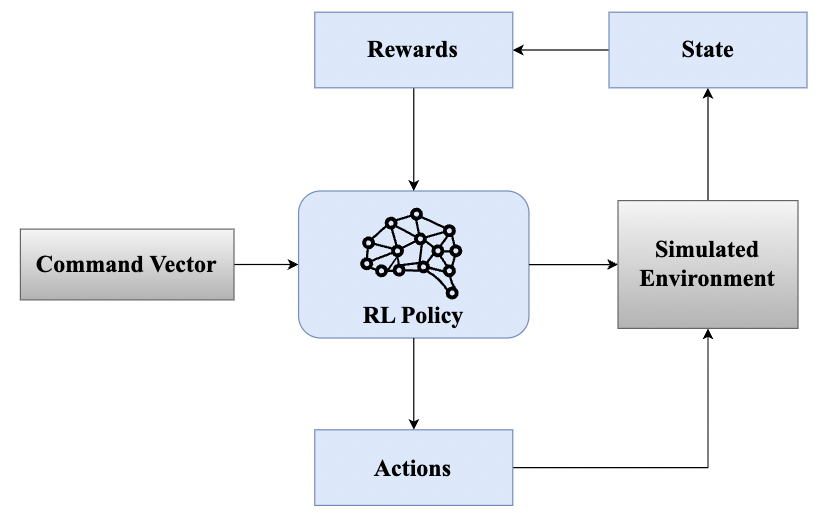}
  \caption{Simulated Training Loop}
  \label{fig:Simulated Training Loop}
\end{figure}
lead to infeasible locomotion scenarios.Moreover, advancements in hardware-in-the-loop testing enables researchers to bridge the gap between simulation and
reality more effectively. By incorporating real-world like sensor
and actuator inputs into the simulation loop, the simulated environment
becomes more closely aligned with the physical robot,
allowing for more accurate training and evaluation of RL policies.\newline
\newline
The choice of algorithm depends on factors such as the
complexity of the task, the desired level of adaptability, and the
available computational resources. Researchers often explore
and compare different algorithms to identify the most suitable
approach for specific legged locomotion scenarios. When a task
is specified in the form of the command vector (policy inputs)
the agent performs the actions which in turn generate rewards
and observations from the environment. Rewards are functions
that help in updating the policy based on actions and states. The
policy generates a trajectory vector of corresponding states, actions,
and rewards given a set of commands.\newline
\newline
The trajectory generated by a RL-policy at time $t$ .

\begin{equation}\label{eq:1}
\tau = \pi_t [(s_0,a_0,r_0),(s_1,a_1,r_1),.........)]
\end{equation}

In equation \ref{eq:1}, $\tau$ represents the trajectory, $\pi_t$ is the policy at time $t$ and $s_t$, $a_t$ and $r_t$ are the state-action-reward pairs that are generated at the respective time-step $T$.\newline
\newline
During the training process, the reinforcement learning (RL) algorithms mentioned in the paper are employed to train the locomotion module. Within the simulated environment, various factors that influence the robot's dynamics, such as terrain friction, the body's center of mass, and the gravity vector in the body frame, are derived and incorporated into the simulation. However, when deploying the trained policy on a real-world robot, certain privileged observations, which were available during training, become unknown to the policy.
To address this challenge, researchers have introduced the utilization of state estimators. These estimators are designed to extract the privileged information from the sensory observations obtained during real-world deployment. By comparing these sensory inputs with the observations made during the simulated training phase, the state estimators can provide comparable results as inputs to the deployed policy. This approach bridges the gap between the simulated training environment and the real-world conditions, enabling the policy to make informed decisions based on the estimated privileged information, similar to how it would during the training.
\begin{figure*}[h!]
  \includegraphics[width=\linewidth]{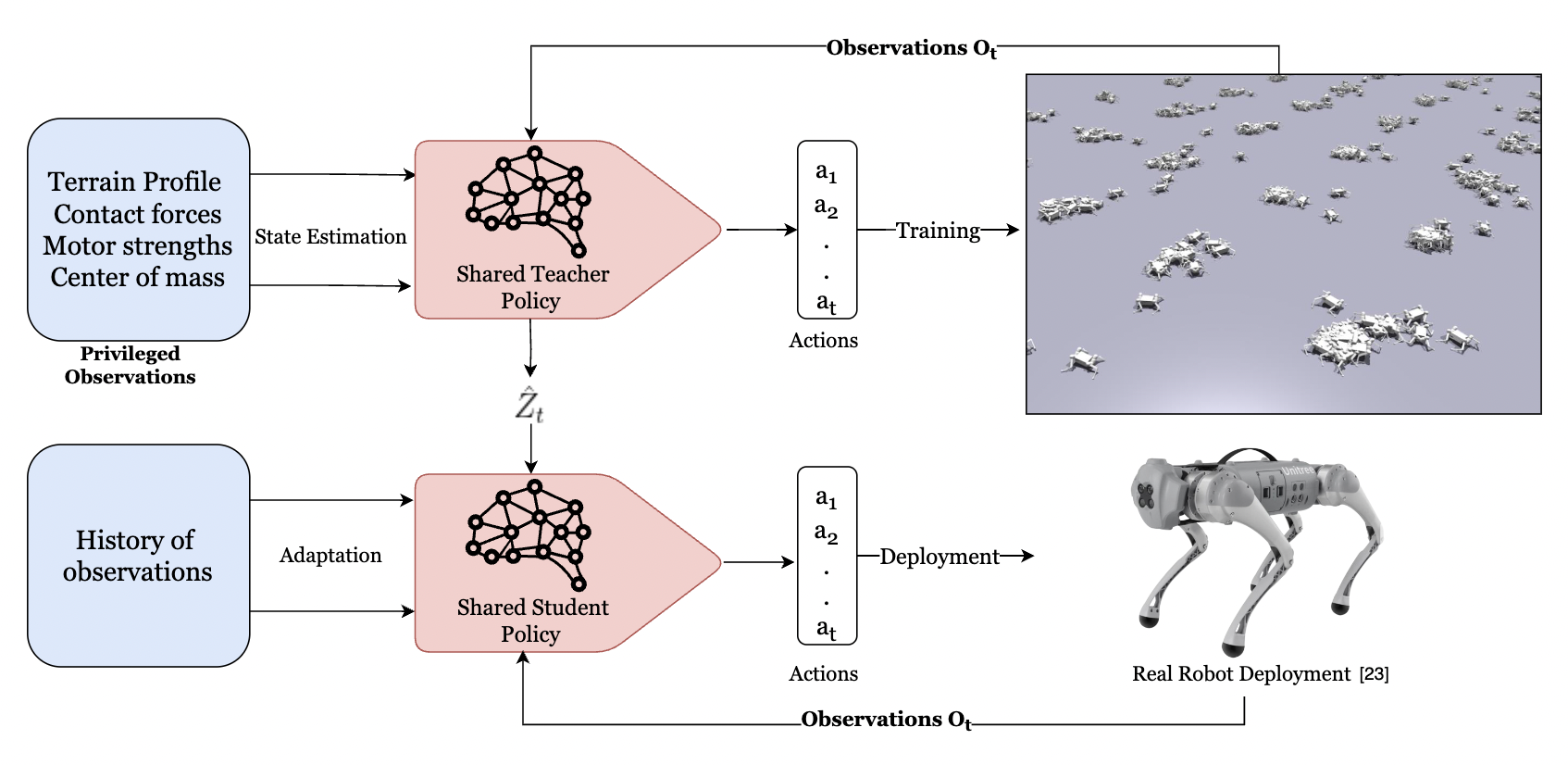}
  \caption{Teacher-Student Framework}
  \label{fig:fig3}
\end{figure*}
\section{Training the locomotion policy}
\textbf{Teacher-student curriculum learning} [9] is an approach employed in machine learning and artificial intelligence that involves training a model using two components: a teacher model and a student model. The learning framework of the teacher-student model has proven to be most suitable for RL-based quadrupedal locomotion. This technique draws inspiration from the way humans acquire knowledge, where a teacher imparts knowledge to a student in a progressive manner. 
The idea behind teacher-student curriculum learning is that by gradually increasing the complexity of the training examples, the student model can learn more effectively and generalize better to unseen data.  In the context of RL-based legged locomotion controllers, the teacher-student model can be used to develop robust locomotion controllers. In this use case, the teacher model can be a pre-trained or expert policy that has already achieved good locomotion control. This teacher model possesses knowledge and expertise in generating effective locomotion strategies for legged robots. The student model, on the other hand, is the RL agent that is being trained to develop its own locomotion controller.\newline
\newline
Joonho Lee [10] and their team were the first ones to build an end-to-end privileged learning-based legged locomotion
model based on [11] which is capable of walking in diverse terrains. Their model employs a teacher policy that trains on privileged information (contact states, contact
forces, and terrain profile) which is further encoded into a latent variable $(lt)$ by passing this information through
an MLP, the policy then gives out the actions based on the states and the privileged information. Once the teacher
policy is trained the model applies the student policy where the policy has no access to the privileged information,
instead, the policy receives the robot’s current state via sensory observations and estimates the latent variable through
the proprioceptive history of observations and imitates the actions of the teacher policy. The proprioceptive history of
observations is recorded every 2 seconds (2/0.02: 100 observations) and is passed through a TCN [12] that estimates
$(lt)$. The model does not use any foot force feedback sensors while deploying, the sensory observations only consist of IMU data and joint angles.
\subsection{Adaptation for RL-based control} 
As mentioned earlier, while training in simulation, the environment’s extrinsic parameters can be provided to the base policy, as these parameters can be computed while training in simulation but are unknown to the policy while deployment on the real robot. RMA (Rapid Motor Adaptation) [13] tackles this by introducing an adaptation module. RMA comprises of two subsystems, namely the base policy and the adaptation module , which collaborate to facilitate real-time adaptation across a wide range of environment configurations. The base policy is trained using reinforcement learning techniques in a simulated setting, where it benefits from privileged information about the environment’s configuration (referred to as vector $et$). This information, encompassing factors such as friction and payload, enables the base policy to effectively adapt to the specific environment. Initially, the environment configuration vector $et$ is encoded into a latent feature space denoted as $zt$, employing an encoder network $(\mu)$. Subsequently, this latent vector $zt$ (referred to as the extrinsic vector) is combined with the current state $(xt)$ and the previous action $(at-1)$ and fed into the base policy to predict the desired joint positions of the robot $(at)$. The policy and the environmental factor encoder $(\mu)$ are jointly trained using reinforcement learning techniques in a simulated environment. The factor encoder gains environmental knowledge from the simulation, the environmental observations are a 17-dimensional vector $et$ that includes motor strength, orientation, friction, and local terrain height, which is then converted into an 8-dimensional extrinsic vector $zt$ via an MLP network of size [256,68].\newline 
\newline
Based on the task that the user wants
to achieve, for example, sprinting on flat ground, omnidirectional velocity tracking, or spinning on uneven surfaces, the user needs to shape the rewards and observations derived from the policy during training. While training
the base policy’s inputs are the states $(st)$, the previous actions $(at)$, and the extrinsic vector $(zt)$, the adaptation module estimates these extrinsic values via supervised learning applied to the history of actions and states. For example the state vector (30-dims) can consist of joint positions, joint velocities, roll and pitch of the trunk, and binarized values of foot contact. As the quadruped robot walks through the environment in simulation, it observes changes in the environment (terrain height, friction, etc.), these observations result in changes to the extrinsic vector based on which the base policy is updated, and further, the behavior is changed. As the model has been trained via supervised learning on the states and action pair history, when encountered with different environmental changes, the state estimator predicts the next actions to be taken
when it receives a similar state and action history. Ji-tendra M et al [14] derive a velocity-conditioned policy that trains the teacher policy using predefined velocity commands; these velocity commands act as experts for the student policy; further, the continuous values between these commands are trained via interpolating and re-sampling the velocity commands at regular intervals.
\section{Learning Curriculum}
To robustify the controller to changes in the dynamics of the environment, we randomize the environments parameters
as well as the task parameters at certain ranges. The expectation from this is that the policy will derive a generalized
solution to these changes (motor strength, friction, proportional gain, derivative gain, center of mass, velocity)
[15]. Generally, training is done in an incremental manner by increasing the complexity of the domain, which may include
increasing the complexity of the terrain, changing friction coefficients, changing the gait parameters, and changing the
task commands to more rigid goals. Every model has its own curriculum strategy to optimize the RL model. The learning curriculum strategy helps the model adapt to disturbances in the real world. The curriculum parameters must be changed according to the robot’s hardware constraints and the task it must achieve, which involves finding appropriate gait parameters in the case of gait-bound models [14,16–19] and finding appropriate environmental and command parameters
in the case of free gait [7,10,12,13,15,20,21,22,24-26] policies. The choice of terrain varies from model to model; previously, there has been
a lot of research on models trained on diverse terrains and models trained on flat ground. Both show properties of traversability in diverse deformable terrains.\newline
\newline
Nikita Rudin et al [22] train their model on multiple parallel terrain types
that include incremental/decremental stairs, rough land, and uneven surfaces with varying friction coefficients, this methodology ensures that the model will perform well on similar terrain types in the real world. Their work employs a curriculum where the difficulty of the terrain and the task is incremented once the agent is able to complete the current task successfully; if the agent fails to adapt to the curriculum, the difficulty is decreased, this method ensures that the agent doesn't face infeasible locomotion conditions, which leads to early termination of training. As information like friction and ground impact cannot be anticipated in the real world, most models apply domain randomization to help the agent get
accustomed to changes in the environment. Simulating soft and granular terrain is a strenuous task, which makes it tough
to train the controller on such terrain types. This problem gave rise to controllers that are trained on flat ground with varying gait parameters [16], these approaches have shown robustness like the models that have been trained on diverse terrains. RMA [13] samples random terrain types while increasing the difficulty of
the task parameters (increasing the linear and angular velocity). \newline
\newline
In Gabriel B. et al [15] the main task is omnidirectional
velocity tracking, they introduce a Grid Adaptive Curriculum strategy, where the liner and angular velocity commands are sampled jointly 
from a probabilistic distribution based on the rewards of the current episode. For example, initially, the velocity commands are given from a smaller range $([-1.0,1.0])$, if the agent receives a positive reward, the range can be further expanded to $([-3.0,3.0])$. The update of the curriculum depends on the rewards generated by the episode, the expansion/contraction of the velocity range is restricted to the grid update region specified by the developer, [15] use $(v_x\in[\pm0.5],\omega_z \in[\pm0.5])$. Jitendra M et al [14] suggest minimizing energy consumption leads to efficient gaits, the selection of gaits at higher or lower speeds affects the energy consumption of the quadruped differently. They hard-code standard gaits and train the controller to minimize energy consumption based on given gait parameters; the complexity of the terrain is increased based on higher frequency commands. \newline
\newline
The training curriculum used by P. Agrawal et al [16] incorporates standardized gaits into their learning curriculum. They claim that diverse terrain travesability can be achieved by switching between gaits (bounding, pacing, trotting, and pronking) for corresponding terrain types. For example, bounding has proven to be the best gait for sandy terrain. For their method, the gait parameters are used as a command input to the policy, where the agent has to maintain these gait parameters while simultaneously maintaining the task (velocity commands) parameters. The curriculum in this case is increasing the complexity of these gait parameters, which include body height, foot-swing height, stance width, and frequency, while simultaneously increasing the complexity of the task parameters. Training via curriculum learning ensures that the controller remains robust even on some terrain types that were not encountered during training. We observed that the controller tries to constantly maintain the stability of the quadruped, even when it is shoved or restricted to move one leg. Recovery is another aspect that the controller learns, where it tries to stabilize itself after drastic foot slippage.\textbf{ While testing P. Agrawal et al [16] on highly diverse and deformable terrains the quadruped recovered from a highly unstable scenario where it was made to descend a $70^{\circ}$ decline on sandy terrain, the quadruped instantaneously performed a swift turning maneuver and stabilized itself. Such examples are highly difficult to simulate and synthetically train. However, based on the given observations, we can assume that given an appropriate training curriculum, the RL conroller will perform better in reality than in simulation in some instances. }  

\subsection{Rewards}
Rewards are the core of any reinforcement learning algorithm; the trajectory generated by the policy has three components [$\pi_\theta(a_t,s_t,r_t)$], amongst which rewards are the deciding factor that drives the algorithm towards the global optimum. The rewards provide feedback to the RL agent, guiding its behavior and shaping the learning process. By designing appropriate reward structures, we can incentivize the RL agent to take actions that maximize the cumulative rewards over time, ultimately leading to the global optimum or desired behavior. The updates made to the policy rely on the rewards that are generated at that time step. The rewards include elements like linear/angular velocity tracking, feet air time, roll pitch, and foot swing duration and also include a set of penalties in case the system performs inefficiently (joint limit violations, joint collisions, failure to maintain height, joint torque). The rewards drive the system towards maintaining smoothness and stability, they vary depending upon the task that the model has to achieve. \newline
\newline
Let us take the example of Tao Chen et al [19] where the task is to make the quadruped cross small gaps (20–30 cm) on the ground using variable and fixed gait constraints, as they train their model on both free and fixed gait patterns, the reward function is modeled accordingly. The reward function generates positive incentives for maintaining the velocity threshold if the body-frame moves along the x-axis successfully. Stability and smoothness are achieved by penalizing the model if desired roll, pitch, and yaw values are not maintained. Smoothness in low-frequency gaits is achieved by minimizing the desired velocity in the reward function.\newline 
\newline
In the case of Proximal Policy Optimization [30], at the end of each episode, the advantage function calculates the discounted sum of rewards and the baseline estimate.

\begin{equation}\label{eq:2}
    L^{PG}(\theta) = \hat{E}_t \left[ \log\pi_{\theta}(a_t|s_t)\hat{A}_t\right]
\end{equation}

Policy gradient methods are a type of reinforcement learning algorithm used to optimize policies (decision-making strategies) for maximizing the expected reward in an environment.
In the equation \ref{eq:2}, $\pi_\theta$ represents the stochastic policy being optimized, where $\theta$ represents the policy's parameters. The policy maps the current state of the environment to a probability distribution over possible actions. $\hat{A}_t$ is an estimator of the advantage function at time step $t$. The advantage function measures how much better or worse a particular action is compared to the average action predicted by the current policy for a given state. The expectation $\hat{E}_t$ denotes an empirical average computed over a finite batch of samples collected by running the policy in the environment. This is done in an iterative process where the algorithm alternates between sampling (collecting data by running the policy) and optimization (updating the policy parameters using the collected data). \newline
\newline
Every model based on the task specified has it's own unique set of rewards. The rewards have to be shaped according to the task that the agent performs in order to successfully train the model. For example, To enhance gait-bound models, P. Agrawal et al [16] have introduced a set of auxiliary and fixed rewards that encourage the controller to maintain gait parameters while accomplishing the given task. The auxiliary rewards are specifically designed to ensure that the gait parameters are maintained, the system is penalized if the specified parameters,, like stance width, foot swing height, and frequency, deviate from the defined range or fail to meet the required criteria. The fixed rewards are agnostic of the auxiliary rewards, they are used to gain stability and smoothness. To avoid early termination of the model due to harsh penalties of the reward function, techniques like reward scaling, curiosity-driven rewards, and intrinsic motivation-based rewards can be used. Additional rewards can be added to optimize power consumption by minimizing the torque generated by the motors while maintaining the desired motion.

\section{Hardware}
With the advent of low-cost qudrupedal robots like Unitree's [23] A1, GO1-edu, and B1, the development of RL-based end-to-end locomotion controllers has become feasible and accessible. These quadrupeds, when procured, come with their own low level controls that run the torque velocity and position modes. As the policy outputs the joint positions, this data can be used as commands for the on-board low level controller. P. Agrawal et al [16] use Lightweight Communications and Marshalling(LCM) as the interface to send commands from the policy to the low level controller and receive sensor data from the robot. This symbiotic relationship between the high-level policy and low-level controller allows for seamless integration of sophisticated control strategies with the hardware capabilities of these robots. 
\begin{figure}
  \centering
  \includegraphics[scale=0.75]{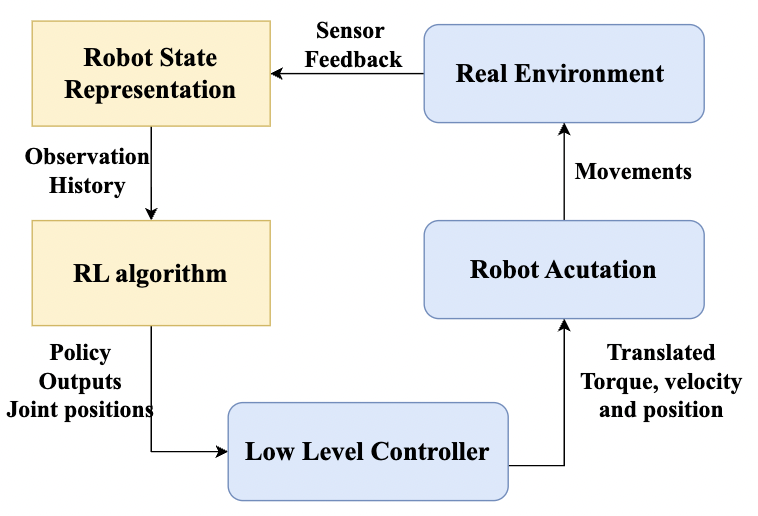}
  \caption{Hardware Control Loop}
  \label{fig:p}
\end{figure}
Inertial Measurement Unit (IMU) sensors are widely used in the context of RL-based controllers for quadrupedal locomotion to provide vital feedback regarding the orientation, acceleration, and angular velocity of the robot. IMU sensors are used to measure the velocity and orientation of robots with respect to magnetic north and the gravity field. Typically, these sensors are accelerometers, gyroscopes, and occasionally magnetometers. These sensors provide data to the trained policy during deployment and represent the current state of the robot. The sensor data is then compared with the proxy data used during training and actions are performed that generated highest rewards during training for comparable type of measurements. Another crucial part of quadrupedal robots' control loops are joint position encoders, which give exact feedback regarding the angles of each joint in their limbs. By measuring the joints' rotational positions, these encoders enable the control system to precisely track the leg configuration of the robot and modify its movements accordingly. 

\begin{longtblr}[
  caption = {Analysis of RL-based locomotion controllers},
]{
  width = \linewidth,
  colspec = {Q[120]Q[84]Q[92]Q[74]Q[62]Q[54]Q[239]},
  hlines,
  vlines,
}
Research Work Title~ & Year and Authors~ & RL Algorithm~ & Gait (Bound/ Unbound)~ & Hardware~ & Sim Tool & Remarks~\\
1) Learning Quadrupedal Locomotion over Challenging Terrain [12]~ & 2020, (J. Lee, J. Hwangbo, L. Wellhausen, V. Koltun, and M. Hutter)~ & Trust Region Policy Optimization (TRPO)~ & Gait free~ & ANYmal~ & RaiSim~ & A blind locomotion model that takes proprioceptive observations from the environment as feedback for the trained student policy for adaptability in diverse terrain.~\\
2) Walk These Ways: Tuning Robot Control for Generalization with Multiplicity of Behavior [16]~ & 2021, (Gabriel B.Margolis , Pulkit Agrawal)~ & Proximal Policy Optimization (PPO)~ & Gait Bound ~ & Unitree’s GO1~ & {Nvidia’s Issac Gym~\\~} & Trained the model with multiple gait parameters(trotting,galloping,pronking) along with various task parameters (velocity, height, stance width, foot swing height). Claim is multiple gaits ensures traversability in diverse terrains.~\\
3) RMA: Rapid Motor Adaptation for Legged Robots [13]~ & 2021, (Ashish Kumar, Zipeng Fu, Deepak Pathak, Jitendra Malik)~ & Proximal Policy Optimization (PPO) altered according to RMA algorithms ~ & Gait Free~ & Unitree’s A1~ & RaiSim Simulator~ & {Presents an adaptation module that is trained to take hardly 2 seconds of observation history from the environment and perform locomotive tasks in diverse terrains.~\\Rewards focused on reducing operational effort bore by the quadruped.~}\\
4) Learning to Walk in Minutes Using Massively Parallel Deep Reinforcement Learning [22]~ & 2022, (Nikita Rudin, David Hoeller, Philipp Reist)~ & Custom Proximal Policy Optimization (PPO)~ & Gait Free (Can adopt any locomotion behavior suitable to the environment)~ & ANYmal B and Unitree’s A1~ & Nvidia’s Issac Gym~ & Utilizes a game inspired curriculum, that increases/decreases the difficulty of the curriculum. Highly reduced policy training time.~\\
5) Rapid Locomotion via Reinforcement Learning [15]~ & 2022(Gabriel B. Margoli, Ge Yang, Kartik Paigwar, Tao Chen, and Pulkit Agrawal)~ & PPO~ & Gait Free~ & MIT’s mini-Cheetah~ & {Nvidia’s Isaac Gym~\\~} & Trained on 4000 parallel simulation models simultaneously. Deploys a ‘Grid Adaptive’ curriculum strategy that expands/contrasts the command vector by increasing/decreasing the difficulty of the task in a selective grid format based on the success/failure of the trial.~\\
6) A Walk in the Park: Learning to Walk in 20 Minutes with Model-Free Reinforcement Learning [24]~ & 2022, (Laura Smith, Ilya Kostrikov , Sergey Levine)~ & Dropout-Q functions for actor critic method~ & {Gait Free~\\~} & Unitree’s A1~ & {MuJoCo~\\~} & Provide a framework that can train the RL controller in comparatively minimal time in both simulated and real environments. Conducts a comparative analysis on damping (Kp) and stiffness (Kd) values for Unitree’s A1 quadrupedal Robot. ~\\
7) Decentralized Policy Learning for Quadruped Robots Multi-Gait Control [17]~ & 2023, (XingCheng Pu, Congde Zhang)~ & Decentralized Multi-Critic Actor Learning~ & Gait Bound~ & Unitree Robotics~ & Nvidia’s Issac Gym  & Harness the principles of limitation learning by tracking the gait-reference motion trajectory and shaping the policy outputs (joint positions) based on it.  Deploy a decentralized leaning strategy that achieves optimality of the control policy by utilizing the multi-agent and multi-task strategy of learning while ensuring optimal convergence between them. ~\\
8)SYNLOCO: Synthesizing Central Pattern Generator and Reinforcement Learning for Quadruped Locomotion [18]~ & 2023, (Xinyu Zhang, Zhiyuan Xiao, Qingrui Zhang, and Wei Pan)~ & Proximal Policy Optimization~ & Gait Bound~ & Unitree’s GO1~ & Nvidia’s Issac Gym~ & A combination of a Central Pattern Generator and Reinforcement Learning. Trained in a two-step method, first by behavior cloning from animal demonstration data and second by training the Reinforcement Learning Feedback Controller based on the data received from the first step.~\\
9) DreamWaQ: Learning Robust Quadrupedal Locomotion with Implicit Terrain Imagination via Deep Reinforcement Learning [35]~ & 2023, (I Made Aswin Nahrendra, Byeongho Yu, Hyun Myung)~ & {Proximal Policy Optimization~\\~} & Gait Free~ & Unitree’s A1~ & {Nvidia’s Issac Gym~\\~} & {Novel locomotion learning framework via asymmetric actor-critic architecture.~\\Context-aided estimator network for body state and environmental context estimation.~}\\
10) Learning Multiple Gaits within Latent Space for Quadruped Robots [32]& {2023, (Jinze Wu, Yufei Xue, Chenkun Qi\\)} & Proximal Policy Optimization (PPO) & Gait Bound & Unitree GO1 & Nvidia’s Issac Gym~ & Train a single policy for multiple gaits that only takes proprioceptive measurements as observations. The training framework is built on an Aysmetric Actor-Critic architecture and uses a GAN discriminator for policy guidance.\\
11) ZSL-RPPO: Zero-Shot Learning for Quadrupedal Locomotion in Challenging Terrains using Recurrent Proximal Policy Optimization [26]~ & {2024, (Yao Zhao, Tao Wu, Yijie Zhu, Xiang Lu, Jun Wang, Haitham Bou-Ammar, Xinyu Zhang, Peng Du~\\)~} & Recurrent Proximal Policy Optimization (RPPO)~ & Gait Free~ & {Unitree A1 and~\\Aliengo~} & {Nvidia’s Issac Gym~\\~} & {Introduced Recurrent Proximal Policy Optimization, an algorithm that is comparatively more robust in nature. They cover a broad range of parameters while applying domain randomization and utilizing exteroceptive measurements during training and deployment.~\\~}\\
12) Learning Quadrupedal High-Speed Running on Uneven Terrain [35]& 2024, (Xinyu Han ,Mingguo Zhao) & Proximal Policy Optimization (PPO) & Gait Bound & Unitree A1 & Webots [36] & Developed a novel model-based joint controller that is highly resistant to external disturbances. A reference trajectory generator and high impedance control stabilize the robot while traversing diverse terrains at high speeds (1.8 m/s). Applies adaptive curriculum learning to prevent overfitting and to increase robustness.~\\
{13) Dexterous Legged Locomotion in Confined 3D Spaces with\\Reinforcement Learning [25]} & 2024, (Zifan Xu, Amir Hossain Raj, Xuesu Xiao, and Peter Stone) & Proximal Policy Optimization (PPO) & Gait Bound & Unitree GO1 & Nvidia’s Isaac Gym~ & A goal-pose-based, end-to-end training approach that is able to navigate through highly dextrous locomotion scenarios.~ Motor skills learned within a restricted parameter range in diverse and challenging spaces. Suggest hybrid dexterity as the most effective method for dextrous manuverability.
\end{longtblr}

\section{Future Scope}
The complete reliance on proprioceptive models is viable only to an extent as we cannot solely depend on the history of observations made by the quadruped, to enhance the robustness of the controller, incorporation of exteroceptive measurements [20,21] is essential. Existing methods utilize the sensory information from various on-board sensors (LIDARs and Depth Cameras) on the quadruped robot, and provide these inputs as additional measurements to the adaptive controller during deployment. The information generated can be used by the model to assess the terrain properties and modulate the footholds and other gait parameters. During training, the teacher policy can use scandots as proxies for the terrain depth measurements from the sensors.This would preemptively alter the gait plan of the quadruped and perform stabilized locomotion. Ananye Agarwal et al [20] deploy this strategy by training in simulation with egocentric measurements from terrain elevation and depth.\newline
\newline
The concept of 'day-dreaming' can come into play, where once the RL-policy is deployed onto the quadruped, it won't just navigate rough terrain but also run an internal training simulation by taking sensory inputs from the environment and use them to improve the learned policy. Philipp Wu et al [31] establish the baseline concept for this via training the quadruped directly in the real world to walk and recover from fall. Their goal is to minimize training time and simulation cost. To protect the hardware from extreme falls and harsh maneuvers, they employ a 'Butterworth Filter' that shields the motor from high-frequency movements. The baseline policy can most likely be used while it is deployed in the real world by letting it train in the back-end by creating different training scenarios and solving them.\newline
\newline
Another intriguing direction is the application of quantum reinforcement learning (QRL) [33] techniques to this domain. QRL is an emerging field that combines principles from quantum computing and reinforcement learning, potentially leveraging quantum computational advantages for solving complex optimization problems more efficiently than classical RL algorithms. The principles of quantum learning can address the exploration-exploitation dilemma in RL. Since an RL model can exist as an orthogonal quantum state, the corresponding eigen-action pairs can also be represented as a superposition state. By applying QRL principles, the rewards of a particular system can be maximized, as the negative rewards generated can be avoided while the system lies in a superposition state.

\section{Conclusion}
This comprehensive study has provided an in-depth exploration of reinforcement learning (RL) based legged locomotion controllers for quadrupedal robots. Throughout the study, we have examined various RL algorithms employed for locomotion control, including Proximal Policy Optimization (PPO), Trust Region Policy Optimization (TRPO), and their variants. We discussed the merits and trade-offs of these algorithms and their suitability for different locomotion scenarios. We explored different curriculum strategies, such as increasing terrain complexity, randomizing environmental parameters, and progressively expanding the command space. These approaches aim to enhance the generalization capabilities of the learned policies, enabling effective navigation across diverse terrains. Identification of techniques like state estimation and domain randomization to bridge the gap between simulated training and real-world deployment (sim-to-real transfer) was also completed. 
Looking ahead, we outlined several promising future research directions, including the incorporation of exteroceptive sensing modalities, the combination of model-based and model-free RL techniques, and the development of online learning capabilities. These avenues hold the potential to further enhance the adaptability, robustness, and real-world performance of RL-based locomotion controllers.
In conclusion, this study provides a comprehensive understanding of the state-of-the-art in RL-based legged locomotion controllers for quadrupedal robots. By consolidating the existing knowledge and highlighting future research directions, we aim to inspire and guide researchers and practitioners in their pursuit of developing more advanced and capable locomotion systems for legged robots.

%Bibliography

\end{document}